 \documentclass[pmlr,twocolumn,10pt]{jmlr} 

\usepackage{float}
\usepackage{placeins} 
\usepackage{capt-of}
\usepackage{booktabs}
\usepackage{soul}
\usepackage{etoolbox} 
\usepackage{needspace}
\usepackage{siunitx}

\jmlrvolume{297}
\jmlryear{2025}
\jmlrworkshop{Machine Learning for Health (ML4H) 2025}

\title[Uncovering Trajectory and Topological Signatures in Sleep Embeddings]{Uncovering Trajectory and Topological Signatures in Multimodal Pediatric Sleep Embeddings}

\author{%
\Name{Scott Ye}\nametag{\thanks{Work initiated during the author’s M.S. paper in Biostatistics at UNC–Chapel Hill and continued in collaboration with the UNC School of Data Science and Society. Short version presented at TS4H Workshop at NeurIPS 2025.}} \Email{scott.ye@ucsf.edu}\\
\addr Department of Radiology, University of California, San Francisco\\[0.35em]
\Name{Harlin Lee} \Email{harlin@unc.edu}\\
\addr School of Data Science and Society, University of North Carolina at Chapel Hill
}

\begin{document}
\maketitle

\begin{abstract}
While generative models have shown promise in pediatric sleep analysis, the latent structure of their multimodal embeddings remains poorly understood. This work investigates \textit{session-wide} diagnostic information contained in the \textit{sequences} of 30-second pediatric PSG epochs embedded by a multimodal masked autoencoder. We test whether augmenting embeddings with (i) PHATE-derived per-epoch coordinates and whole-night movement descriptors, (ii) persistent homology summaries of the embedding cloud, and (iii) EHR yields task-relevant signals. Simple linear and MLP models, chosen for interpretability rather than state-of-the-art performance, show that geometric, topological, and clinical features each provide complementary gains.
For binary predictions, feature importance is task-dependent, and more expressive late-fusion models generally perform better, with AUPRC improving 0.26→0.34 for desaturation, 0.31→0.48 for EEG arousal, 0.09→0.22 for hypopnea, and 0.05→0.14 for apnea. We also report Brier score and Expected Calibration Error, where the full fusion model yields the best calibration across all four binary tasks. Our study reveals that latent geometry/topology and EHR offer complementary, interpretable signals beyond embeddings, improving calibration and robustness under extreme imbalance.
\end{abstract}


\paragraph*{Data and Code Availability}
The Nationwide Children’s Hospital Sleep DataBank is available at NSRR~\citep{zhang2018national} and Physionet~\citep{goldberger2000physiobank}. All preprocessing, feature extraction, and model-training scripts are available at \href{https://github.com/scottye009/PedSleep-TTA}{\texttt{https://github.com/scottye009/PedSleep-TTA}}.

\paragraph*{Institutional Review Board (IRB)}
This work analyzes a publicly released, de-identified dataset and does not constitute human subjects research. Therefore, IRB approval is not required.

\section{Introduction}
Sleep disorders in the pediatric population significantly impact developmental health, cognition, behavior, and cardiometabolic outcomes~\citep{ANDERS19979, Marcus2012, american2007aasm}. 
Understanding pediatric sleep presents unique challenges compared to adults: respiratory events are shorter and subtler, arousals and hypopneas are harder to detect, and clinical scoring criteria differ~\citep{american2007aasm, berry2012rules}. 

\begin{figure*}[ht]
  \centering
  \includegraphics[width=\linewidth]{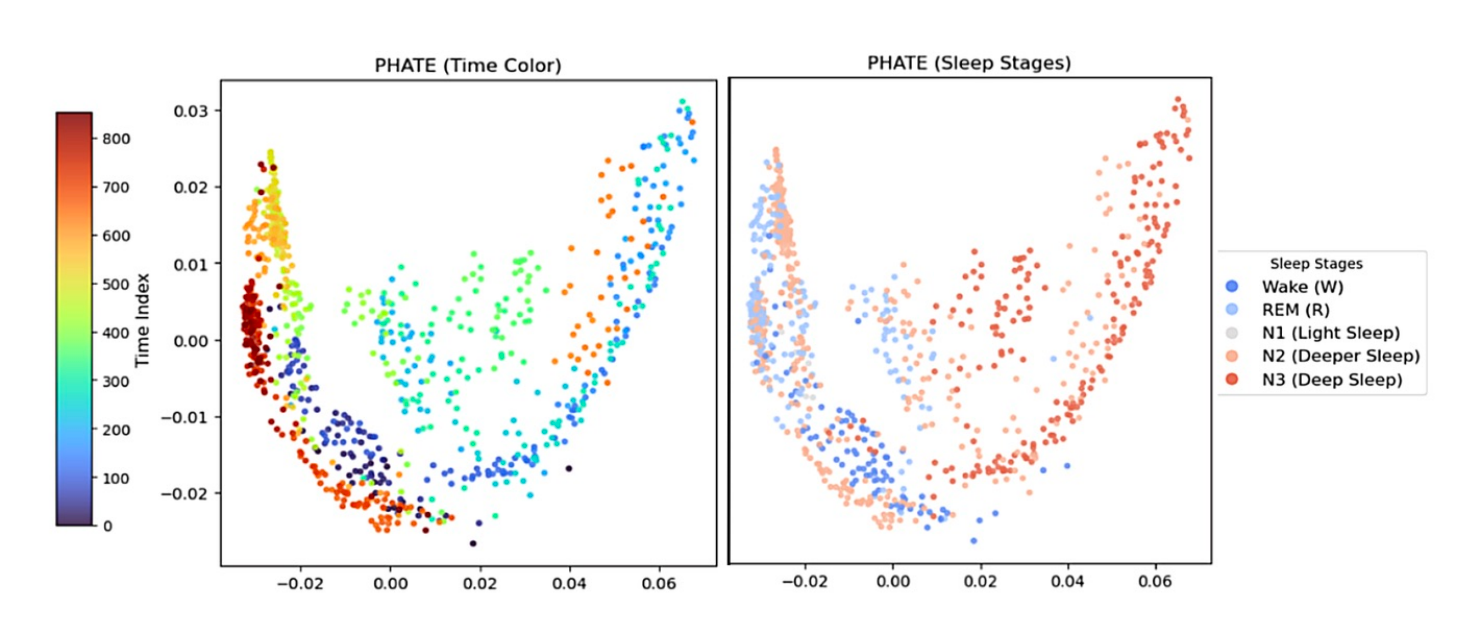}
  \caption{Parallel PHATE views for one sleep session: left colored by epoch index (=time of night), right by sleep stages. Each point is multimodal embedding of 30 seconds of sleep (=1 epoch). The embedding model was trained without access to the epoch index information. Still, the 2-D diffusion map reveals a smooth, time-ordered trajectory whose regions align with expert staging.}
  \label{fig:a1}
\end{figure*}

Overnight polysomnography (PSG) provides rich multimodal recordings for clinical diagnosis, and most recently, for generative learning~\citep{pandey2024}. 
These models are shown to identify pediatric sleep stages and events, but they raise a deeper question: 
what do they actually encode? 
Much less attention has been paid to understanding whether \textit{their latent structure captures higher-level information about disease burden, sleep continuity, or clinical severity}. This gap is particularly important in pediatrics, where small differences in scoring rules and the lower prevalence of events that vary between demographic strata (Appendix Tables~\ref{tab:app-prevalence-part1} and~\ref{tab:app-prevalence-part2}) can lead to substantial differences in diagnosis and treatment decisions~\citep{Marcus2012, berry2012rules}.


We present a motivating example in Figure~\ref{fig:a1}.
It visualizes PedSleepMAE~\citep{pandey2024} embeddings, which are fixed, multimodal representations learned generatively from raw pediatric PSG  via a masked autoencoder \citep{he2022masked}. 
PedSleepMAE was trained by treating every 30 seconds of PSG as an independent sample, i.e.\ reconstructing masked signals without knowing time of night or patient identity. Yet Figure~\ref{fig:a1} reveals that the embeddings captured time-dependent structure despite not knowing it explicitly in training. PHATE~\citep{Moon2019} maps each night to a smooth, time‐ordered path whose geometry matches expert stages: lighter stages at the entrance, N3 near the center, and REM along peripheral arcs. Across sessions we observe consistent curvature, drift, and occasional bifurcations that align with canonical sleep progressions. 

This motivates our novel research question of investigating the \emph{session-wide diagnostic information} contained in the \emph{sequences of multimodal generative embeddings}. To answer this question, we ask whether the embeddings' (a) latent trajectory information, (b) topological shape, and (c) augmentation with EHR can (i) reflect disease burden across AHI strata and (ii) improve detection of apnea, hypopnea, desaturation, EEG arousal, and five sleep stages.

Concretely, we map per-epoch PedSleepMAE embeddings to 2-D PHATE and treat each study as a smooth latent trajectory, yielding (i) \emph{trajectory-local} per-epoch coordinates/derivatives and (ii) \emph{trajectory-global} movement/fragmentation summaries. 
In parallel, we compute persistent homology directly on the original 7{,}680-D embedding cloud and summarize $H_0$ and $H_1$ characteristics with a compact and stable panel. Routine EHR (age/sex and common pediatric comorbidities) provides low-overhead clinical context that can reduce confounding and improve generalization when fused with signal features. 

Unlike prior work that primarily treats embeddings as inputs for classification, 
we study their session-wide trajectory and topological signatures. 
Our contributions are:
\begin{itemize}
    \item \textbf{Trajectory analysis:} We perform session-wide study of pediatric PSG embeddings, moving beyond per-epoch classification and visualization to analyze latent \emph{trajectories} that capture time-dependent sleep dynamics.
    \item \textbf{Topological characterization:} We apply manifold learning and persistent homology directly to high-dimensional PedSleepMAE embeddings, yielding stable, compact signatures of sleep continuity and fragmentation.
    \item \textbf{Clinical fusion:} We show that augmenting embeddings with geometric features and routine EHR covariates improves generalization across AHI strata and enhances sleep event detection.
\end{itemize}
Together, these results suggest a new approach to interpreting generative models in sleep medicine.

\section{Background and Related Work}

\paragraph{Terminology.} \textit{Session} is a sleep study or PSG. \textit{Epoch} is 30-seconds of sleep, which is a clinical term unrelated to training iterations in machine learning. \textit{Subject} is a patient with PSG(s).

\paragraph{Physiological signals.}
Overnight PSG typically includes EEG/EOG/EMG, airflow, thoracoabdominal effort, ECG, and SpO$_2$ at 100--500\,Hz (channels vary by site). EEG/EOG/EMG support sleep staging; airflow/effort capture apneas/hypopneas; SpO$_2$ reflects desaturation burden. Our labels follow this physiology: staging from EEG/EOG/EMG, respiratory events from airflow/effort, desaturation from SpO$_2$, and arousals from EEG conventions.

\paragraph{Deep learning in pediatric sleep.}
Most studies focus on per-epoch classification of sleep stages or respiratory events~\citep{supratak2017deepsleepnet, phan2021xsleepnet,lee2022automatic} with the aim of automating resource-intensive manual labeling. With the rise of AI, increasing number of works explore self-supervised learning or generative modeling~\citep{banville2021uncovering}, including in foundation models~\citep{pmlr-v235-thapa24a} and in pediatric sleep~\citep{pandey2024}. Our work analyzes the geometry and topology of embeddings generated by such models.

\paragraph{Manifold learning in physiological signals.}
Manifold learning is widely used to visualize high-dimensional trajectories~\citep{Becht2019}. PHATE’s diffusion geometry preserves local neighborhoods while maintaining global progression and denoises noisy biological measurements, making it suitable for sleep dynamics~\citep{kuchroo2020}. Prior PSG work more often models raw or time–frequency inputs with sequence architectures, e.g., SleepTransformer~\citep{phan2022sleeptransformer}. Our approach goes beyond visualization (e.g.,~\citep{banville2021uncovering}) and leverages \emph{trajectory geometry in representation space} to summarize whole-night dynamics.

PHATE preserved both local continuity and global progression better than UMAP in Appendix~\ref{app:umap} Figure~\ref{fig:app-phate-umap}; PHATE formed a smooth temporal manifold while UMAP fragmented it into disconnected clusters. In an ablation study (Appendix~\ref{app:umap} Table~\ref{tab:app-umap}), swapping PHATE with UMAP reduced AUPRC on three of four binary tasks while slightly hurting sleep scoring F1.

\paragraph{Topological data analysis (TDA).}
Persistent homology offers stable vectorizations that capture multiscale loop/cluster structure for learning~\citep{bubenik2015statistical, adams2017persistence, atienza2018stability}. Pediatric sleep EEGs have related such structure to respiratory burden and desaturation~\citep{sathyanarayana2025topological}. We extend this to latent spaces learned from multimodal PSG, adding complementary structure beyond pointwise embeddings.

\begin{table*}[t]
  \centering
  \caption{EHR, trajectory, and TDA-related features considered in this study.}
  \vspace{0.5em}
  \label{tab:branch-inventory}
  \begin{tabular}{@{}p{3.2cm}p{2.0cm}p{1.0cm}p{9.8cm}@{}}
     \toprule
    Branch & Scope & Dim & Contents \\
    \midrule
    EHR--Demographics & per-session & 11 &
    age, gender(3), race(6), ethnicity(1). \\
    EHR--Comorbidities & per-session & 12 &
    asthma, obesity, diabetes, hypertension, depression, anxiety, ADHD,
    seizure disorder, GERD, cerebral palsy, autism, developmental delay. \\
    Trajectory-local \hspace{1cm} (2-D PHATE) & per-epoch & 6 &
    delta distance($\delta_t$), cumulative distance($\mathrm{cum}_t$), turn angle($\mathrm{turn}_t$), curvature($\mathrm{curv}_t$), distance from start, segment ID (PELT). \\
    Trajectory-global  \hspace{1cm} (2-D PHATE) & per-session & 6 &
    $\mathrm{mean}(\delta_t)$, $\max(\delta_t)$, $\mathrm{mean}(\mathrm{turn}_t)$,
    directional entropy, tortuosity, $n_{\text{segments}}$. \\
    Topological features (7,680-D embedding) & per-session & 6 &
    $H_0$ sum persistence, $H_0$ number of bars, $H_1$ number of bars, $H_1$ max persistence, 
    Betti--$L^2$, $H_1$/$H_0$ persistence ratio. \\
    \bottomrule
  \end{tabular}
\end{table*}

\paragraph{Electronic Health Records (EHR).}
Combining signal representations with structured EHR via late fusion is a common and effective pattern in clinical prediction~\citep{10.1093, huang2020multimodal}. Our results align with this pattern: EHR helps for the rarest outcome (apnea), while trajectory and topology features contribute more for desaturation, hypopnea, and EEG arousal.

\section{Methods}
Our methodological pipeline is designed to test whether multimodal pediatric sleep embeddings encode clinically meaningful structure. After describing the PSG and EHR dataset, we outline the derivation of geometric and topological features, perform an AHI-stratified feature analysis to examine their relationship with disease severity, and then evaluate their predictive ability through late-fusion models.

\subsection{PSG and EHR Data} \label{sec:data}
We use the pediatric overnight polysomnography (PSG) from Nationwide Children’s Hospital Sleep DataBank (NCHSDB)~\citep{Lee2022}. The analysis set includes 2{,}522 PSGs (2,379 unique subjects), each identified by a (subject ID, session ID) pair. Recordings are divided into consecutive 30-second epochs. Each epoch is represented by a 7{,}680-D PedSleepMAE embedding (120$\times$64) learned generatively from raw PSG channels \citep{pandey2024}. Modalities considered by PedSleepMAE include 7-channel EEG, 2-channel EOG, EMG, snoring, respiratory effort, airflow, oxygen saturation, and CO2 level.  Labels for sleep stage, apnea, hypopnea, desaturation, and EEG arousal align one-to-one with each epoch. We use subject-wise, stratified splits (70/10/20\% train/val/test) per label and 5-fold subject-wise cross-validation.

Structured EHR from NCHSDB are also linked to each subject ID. Routine EHR provides clinical context that can reduce confounding and improve generalization when fused with PSG features. Therefore, we include a demographic and comorbidity set in our analysis (Table~\ref{tab:branch-inventory}). All EHR variables are session-level, as subjects with multiple PSGs may have different values for, e.g., age.

\paragraph{Subgroup prevalence.}
We compute per-epoch positive rates by age group, sex, and race on the study population and report them for all labels in Appendix~\ref{app:prevalence} Tables~\ref{tab:app-prevalence-part1} and~\ref{tab:app-prevalence-part2}.

\subsection{Feature Sets}
Our features mirror the ablation order: per-epoch PedSleepMAE embeddings as baselines, EHR (Sec.~\ref{sec:data}), then (i) PHATE-based trajectory features and (ii) topological descriptors.

\paragraph{PHATE trajectory features.}
2-D PHATE is fit on training sessions and applied out of sample to validation/test. 
We use (a) \emph{trajectory-local} per-epoch coordinates/derivatives and (b) \emph{trajectory-global} session summaries of movement/fragmentation: mean and max inter-epoch delta distance, mean turning angle, directional entropy of turns, tortuosity (path-length vs.\ end-to-end), and a change-point count on the step-length series using \textsc{ruptures} with PELT (Pruned Exact Linear Time) \citep{truong2020selective, Killick2012PELT}. Session-level quantities are broadcast to all epochs of that session.

\begin{minipage}[t]{0.48\textwidth}
\subsubsection*{PHATE trajectory quantities}
Let $p_t = [p_t^x, p_t^y]\in\mathbb{R}^2$ denote PHATE coordinates at epoch $t$. $t$ goes from 1 to $T$ in a given session. 
\begin{align*}
\delta_t &= \lVert p_t - p_{t-1}\rVert_2 \\
\mathrm{cum}_t &= \sum_{i=2}^t \delta_i \\
\theta_t &= \operatorname{atan2}(p_t^y-p_{t-1}^y,\, p_t^x-p_{t-1}^x) \\
\mathrm{turn}_t &= \theta_t-\theta_{t-1} \text{~in~} (-\pi, \pi] \\
\mathrm{curv}_t &= \tfrac{|\mathrm{turn}_t|}{\delta_t+\varepsilon} \\
\mathrm{dist\_start}_t &= \lVert p_t - p_1\rVert_2 \\
\mathrm{dir\_entropy} &= -\sum_b \hat p_b \log \hat p_b \\
\mathrm{tortuosity} &= \tfrac{\sum_t \delta_t}{\lVert p_T-p_1\rVert_2+\varepsilon} \\
n_{\text{segments}} &= \#\{\text{PELT changepoints on }\delta_t\}
\end{align*}
\end{minipage}
\hfill
\begin{minipage}[t]{0.48\textwidth}
\vspace{0pt}
\end{minipage}

\paragraph{Topological features.}

We extracted a wide
panel of persistence-derived statistics on the original 7{,}680-D PedSleepMAE point cloud. For stability and interpretability in the late-fusion
model, we retained six robust statistics as the TDA features: H0 sum persistence, $H_0$ number of bars, $H_1$ number of bars, $H_1$ max persistence, Betti–1 $L^2$ norm, and the $H_1$/$H_0$ lifetime ratio. These capture cluster spread/fragmentation ($H_0$), loop prevalence/strength ($H_1$/Betti–1 energy), and loop-vs-cluster balance, producing stable, fixed-length vectors for learning \citep{bubenik2015statistical,adams2017persistence,atienza2018stability}.

To provide intuition: $H_0$ statistics measure how dispersed or fragmented the embedding cloud is, reflecting continuity of sleep trajectories. $H_1$ descriptors capture the presence and persistence of loops in the latent space, which corresponds to recurrent cycles. Ratios such as $H_1/H_0$ summarize the balance between fragmentation and cyclic structure.

\vspace{0.8em}
\begin{minipage}[htbp]{0.48\textwidth}
\subsubsection*{Topological descriptors}
Let $\mathcal{X}=\{x_i\}$ be the 7{,}680-D embedding cloud; compute
Vietoris–Rips persistence with $H_0$ and $H_1$ barcodes lifetimes
$\{\ell^{(0)}_j\}$, $\{\ell^{(1)}_k\}$.
\begin{align*}
H_0\_\mathrm{sum\_pers} &= \sum_j \ell^{(0)}_j \\
H_0\_\mathrm{n\_bars} &= \#\{\ell^{(0)}_j>0\} \\
H_1\_\mathrm{n\_bars} &= \#\{\ell^{(1)}_k>0\} \\
H_1\_\mathrm{max\_pers} &= \max_k \ell^{(1)}_k \\
\text{Betti--}L^2 &= \lVert \beta_1(r)\rVert_2 \\
\text{ratio\_sum\_$H_1$\_$H_0$} &= \tfrac{\sum_k \ell^{(1)}_k}{\sum_j \ell^{(0)}_j+\varepsilon}
\end{align*}
\end{minipage}

\subsection{AHI-stratified feature analysis}
The apnea–hypopnea index (AHI) is a standard measure of sleep-disordered breathing, 
defined as the average number of apneas and hypopneas per hour of sleep 
\citep{american2007aasm, berry2012rules}. Higher AHI values indicate more severe 
sleep-disordered breathing. We grouped sessions by pediatric AHI thresholds into 
healthy ($<$1), mild (1--5), moderate (5--10), and severe ($\ge$10), following 
commonly used pediatric criteria \citep{Marcus2012}. 

For each session-level candidate we ran a Kruskal--Wallis omnibus test \citep{kruskal1952}, 
Dunn post-hoc comparisons \citep{dunn1964} with Holm correction \citep{holm1979}, 
reported Cliff's $\delta$ as an effect size \citep{cliff1996}, and visualized 
box/ECDF/KDE distributions with adjusted $q$ values. 

Because AHI is defined per session, the screen applied only to trajectory-global PHATE features 
and to TDA summaries. EHR features were included as a pre-specified confounder block and also 
subjected to the same testing for completeness, but were not used for feature selection 
in order to avoid label leakage and to preserve a stable set of clinical covariates across all tasks. All association tests are observational; reported effects are correlational and do not imply causality.

\begin{figure*}[t]
  \centering
  \includegraphics[width=\linewidth]{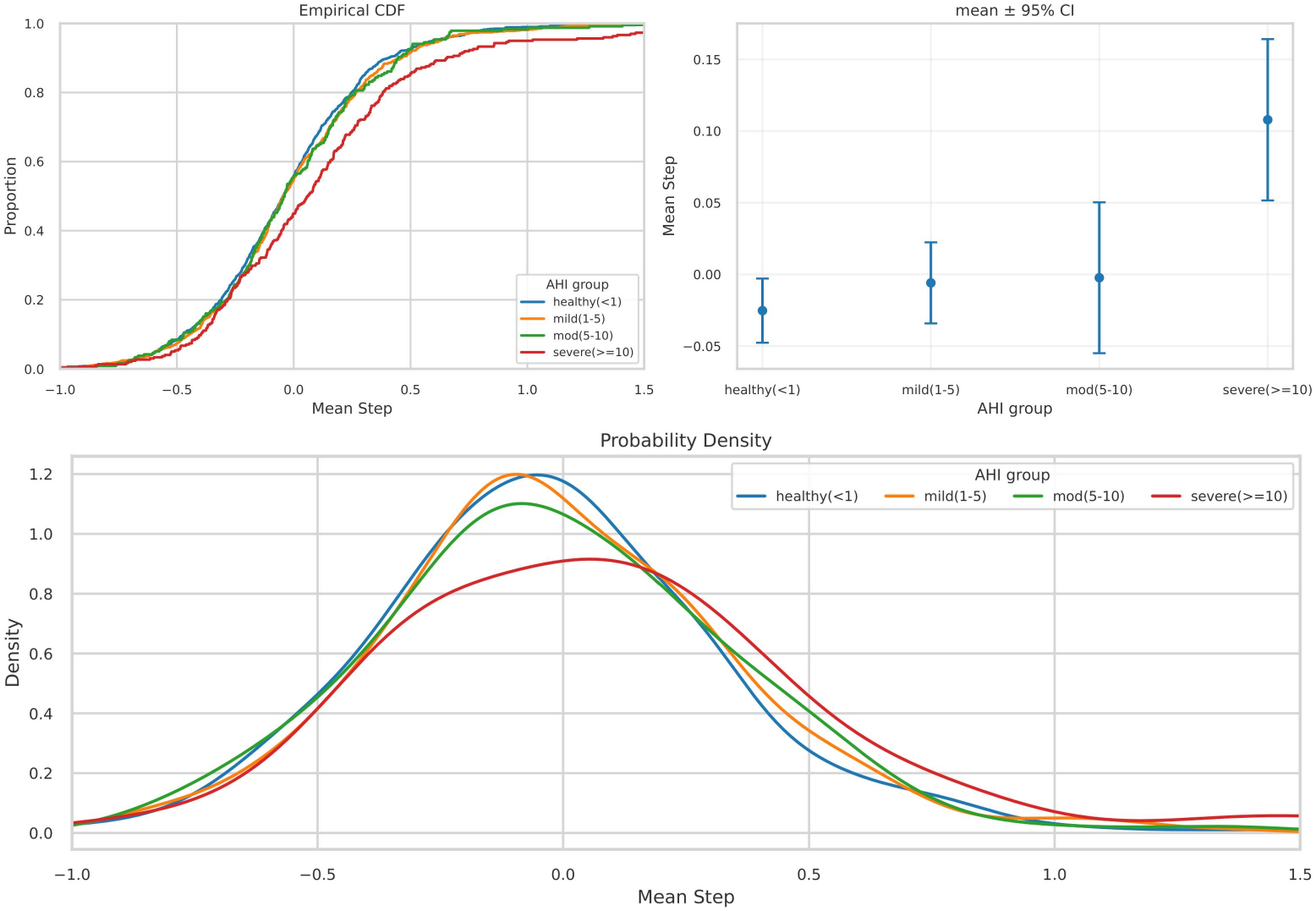}
  \caption{AHI associations for a representative metric ($\mathrm{mean}(\delta_t)$). 
  Top-left: ECDF; top-right: mean$\pm$95\% CI; bottom: KDE density. 
  Groups: healthy ($<$1), mild (1--5), moderate (5--10), severe ($\ge$10).}
  \label{fig:2}
\end{figure*}
\begin{figure*}[th]
  \centering
  \includegraphics[width=0.99\linewidth]{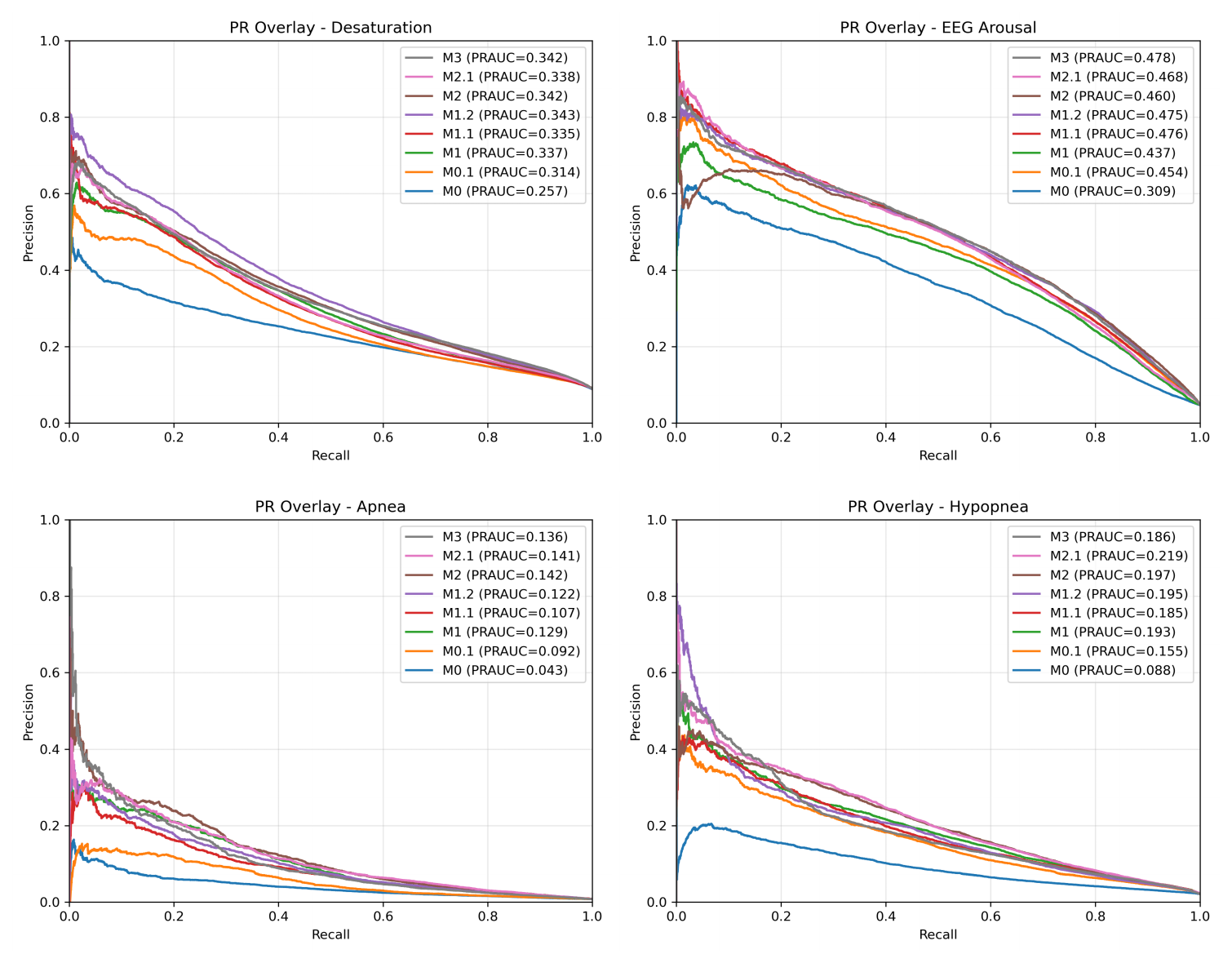}
  \caption{Precision–recall curves for the four binary tasks on the test set. 
  Each panel overlays all ablation models (M0–M3), with legend reporting AUPRC.}
  \label{fig:3}
\end{figure*}

\subsection{Diagnostic models}
We benchmarked a family of late-fusion classifiers on identical train/val/test splits. 
Given the extreme class imbalance (Table~\ref{tab:prevalence}), we used class-weighted and focal losses.
Our goal was not to maximize raw performance with deep or complex architectures, but to use deliberately simple models, linear probes and shallow MLPs, to isolate and interpret the incremental value of each feature family. See Appendix~\ref{app:implementation} for implementation details.

\paragraph{Model variants.}
Our baselines, \textbf{M0 (Emb-linear)} applies a single linear layer directly to each 7{,}680-D PedSleepMAE embedding, setting a lower bound on representation quality \citep{pandey2024}. \textbf{M0.1 (Emb-MLP)} increases capacity by passing embeddings through a shallow encoder MLP.  

\textbf{M1 (Emb+EHR)} uses a two-branch late-fusion MLP: embeddings and structured EHR features (23-D; Table~\ref{tab:branch-inventory}) are encoded separately and concatenated.  
\textbf{M1.1 (Emb+PHATE)} fuses embeddings with both trajectory-local and trajectory-global PHATE features.  
\textbf{M1.2 (Emb+TDA)} fuses embeddings with persistent-homology descriptors from the embedding cloud.  
\textbf{M2 (Emb+EHR+PHATE)} adds both PHATE branches to M1.  
\textbf{M2.1 (Emb+EHR+TDA)} adds TDA instead.  
\textbf{M3 (Emb+EHR+PHATE+TDA)} incorporates all five branches in a full late-fusion MLP. Ablations are ordered for interpretability and deployment. This design allows us to isolate the incremental diagnostic value of clinical context (EHR), trajectory geometry (PHATE), and topology (TDA).

\section{Results}
We present results in two stages. First, we perform an AHI-stratified analysis to examine how geometric, topological, and EHR features vary across clinical severity levels, assessing whether latent structure encodes disease-related information. We then evaluate the predictive performance of late-fusion models across multiple diagnostic tasks, comparing ablation models to determine the individual and complementary contributions of each feature family.
\subsection{Clinical Association with AHI} \label{sec:ahi}
Trajectory movement, topology, and routine EHR features all co‐vary with AHI. 
See Appendix~\ref{sec:app_ahi} for the full omnibus and contrast tables. 
Permutation Kruskal--Wallis omnibus tests (Table~\ref{tab:ahi-omnibus}) 
remained significant for all six topological descriptors and multiple PHATE movement metrics. Severe nights exhibited reduced topological richness 
(fewer connected components and loops, lower Betti energies), 
larger average and more variable manifold steps, 
and higher $H_1$\_max\_pers. 
Healthy nights showed the opposite pattern, with greater topological diversity 
and smoother, less variable trajectories. Pairwise contrasts (Table~\ref{tab:ahi-contrasts}) 
reinforce these findings, showing statistically significant effect sizes in the expected directions 
(e.g., negative $\Delta$ for topological counts in severe cases, positive $\Delta$ for movement variance). The pattern indicates that the metrics are not just differentiating extremes, but capturing 
a graded trajectory of disease burden.

Figure~\ref{fig:2} illustrates these trends for a representative PHATE-derived movement metric ($\mathrm{mean}(\delta_t)$). 
The empirical CDF shows a general rightward shift as severity increases: healthy nights cluster at lower mean steps, while severe nights accumulate probability mass later. 
The mean $\pm$95\% CI plot shows broad overlap among healthy, mild, and moderate groups, with clearer separation for the severe group. 
The KDE density estimate reinforces this pattern, showing both a rightward shift and a broader distribution for severe cases, including a heavier right tail that reflects greater heterogeneity. 
These views suggest that higher AHI severity is associated with larger and more variable movement steps, with the strongest distinction seen between severe and non-severe groups, consistent with the omnibus and contrast test results.

Beyond movement and topology, our extended screen revealed that several EHR variables also stratify by AHI. 
Omnibus tests (Table~\ref{tab:ahi-ehr-omnibus}) flagged obesity, diabetes, hypertension, anxiety, male sex, 
and race (Black, White) among the strongest signals ($q < 0.005$), with additional demographic and comorbidity 
features (e.g., age, depression/mood disorder, GERD) passing at $q < 0.05$. 

Contrasts (Table~\ref{tab:ehr-contrasts}) clarify these effects. We omit the median-difference effect 
($\Delta = \mathrm{median}(\text{group}) - \mathrm{median}(\text{others})$) 
in the table because values were small; nonetheless, several features show 
significant distributional shifts as captured by positive Cliff’s $\delta$. Healthy cases were enriched for female sex, lower 
cardiometabolic burden, and White race, while severe cases showed higher prevalence of hypertension and Black 
race (with White race underrepresented). These associations suggest that routine demographic and comorbidity 
indicators carry complementary disease-burden information alongside PSG-derived descriptors, and reinforce 
their role as a stable context/confounder block for predictive modeling.

\subsection{Predictive Performance}

We report AUPRC as the primary discrimination metric for highly imbalanced clinical tasks~\citep{saito2015precision}, complemented by balanced accuracy, macro–F1, and ROC–AUC (Tables in Appendix~\ref{app:results}). To assess calibration, which is critical for clinical interpretability, we additionally report the Brier score~\citep{glenn1950verification} and Expected Calibration Error (ECE)~\citep{naeini2015obtaining}. Lower Brier/ECE values indicate better-calibrated probability estimates. Full cross-validated metrics are summarized in Appendix~\ref{app:metrics}: Table~\ref{tab:staging} (staging), Table~\ref{tab:desat} (desaturation), Table~\ref{tab:eeg} (EEG arousal), Table~\ref{tab:apnea} (apnea), and Table~\ref{tab:hypop} (hypopnea). Figure~\ref{fig:3} overlays PR curves for the binary tasks, illustrating clear separation from the linear-probe baseline once contextual branches are added.

\paragraph{Sleep staging (5-class).}
Performance improves steadily across the ablation ladder: M0 (linear baseline) reaches F1 0.655±0.001; M0.1 (MLP baseline) improves to 0.664±0.004. Adding context gives consistent lifts, with M2 (Emb+EHR+PHATE) at 0.679±0.003 and the full M3 at 0.678±0.004. Balanced accuracy follows the same trend (M3: 0.762±0.004).

\paragraph{Desaturation.}
AUPRC increases from 0.257±0.022 (M0) to 0.343±0.028 with M1.2 (Emb+TDA). The full M3 remains competitive (0.342±0.022) and yields the best calibration (Brier 0.138, ECE 0.183). Improvements are mainly driven by the addition of PHATE and topological features, which agrees with Sec.~\ref{sec:ahi} where variable sleep trajectories differentiated AHI strata. The lower Brier/ECE values indicate not only better discrimination but also more reliable probability estimates for clinical interpretation.

\paragraph{EEG arousal.}
Trajectory features dominate this task. AUPRC rises from 0.309±0.013 (M0) to 0.476±0.015 (M1.1), and the full M3 performs best overall at 0.478±0.014 with the strongest calibration (Brier 0.084, ECE 0.130). These results reinforce the role of trajectory features, consistent with their AHI monotone shifts.

\paragraph{Apnea.}
The rarest label benefits the most from the clinical context. AUPRC increases from 0.043±0.013 (M0) to 0.142±0.029 with M2 (Emb+EHR+PHATE). Although M3’s discrimination is comparable (0.136±0.036), it exhibits the lowest Brier (0.014) and ECE (0.016). These gains arise mainly from demographic and comorbidity features that encode baseline disease burden (hypertension, obesity, and race) which in Sec.~\ref{sec:ahi} were significantly associated with AHI severity. These patterns suggest the importance of routine EHR context when predicting rare but clinically consequential events.

\paragraph{Hypopnea.}
Signals are complementary: AUPRC moves from 0.088±0.011 (M0) to 0.219±0.020 with M2.1 (Emb+EHR+TDA). The full M3 provides the best calibration (Brier 0.067, ECE 0.103) while remaining competitive in AUPRC (0.186±0.022). Improvements here reflect the joint effect of topological and clinical context: topology contributing sensitivity to subtle respiratory variability, and EHR stabilizing predictions across subjects with differing baseline risk. The trend aligns with Sec.~\ref{sec:ahi}, where both TDA and EHR features showed graded associations with AHI.

\section{Discussion}
Our results support our hypothesis that latent geometry and topology encode physiologically meaningful structure in pediatric sleep.
Importantly, our results should not be interpreted as incremental gains in classification accuracy. Rather, they show that generative embeddings contain rich session-wide structure: by analyzing trajectories and topology, we uncover interpretable signatures of sleep fragmentation and disease severity that complement conventional machine learning models. The present approach provides one interpretive tool for inspecting learned embedding spaces and generating research hypotheses. It is therefore not a clinical decision system and requires further study on external datasets and workflow integration.

\paragraph{No single modality dominates.} Across predictive tasks, no single modality dominates. EHR provides the most value for the rarest and most clinically anchored label (apnea, best AUPRC with Emb+EHR+PHATE), PHATE trajectory features are most impactful for EEG arousals, and TDA is strongest for desaturations. Hypopnea benefits from combining EHR with TDA (best AUPRC with Emb+EHR+TDA). Notably, the full fusion models (M3) consistently delivers the best calibration (lowest Brier/ECE) across all four binaries, even when another ablation narrowly wins AUPRC.

\paragraph{Why not apply PHATE/TDA directly to raw signals?}
A natural question is why our analyses focus on learned embeddings rather than the original PSG waveforms. 

Raw epochs are extremely high-dimensional: a single 30-second segment spans tens of thousands of samples across 16 multimodal channels with distinct noise characteristics. In this space, computation is prohibitive and distance metrics are dominated by artifacts rather than physiology, making the geometry unreliable for manifold learning or TDA.

PedSleepMAE embeddings instead provide compact, multimodal representations trained to reconstruct missing segments, which enforces denoising and cross-channel learning. Distances in this latent space reflect meaningful similarity and yield neighborhoods that are stable, smooth, and well-suited for our analysis.

Moreover, analysis in the embedding space aligns with what many downstream classifiers ``see,'' providing a way to peer into the model’s feature space. While our study focuses on PedSleepMAE, the same approach can be applied to embeddings from any model, making our framework a general tool for interpreting how machine learning models re-structure complex sleep signals.

\paragraph{Limitations and future work.}
Our study has several limitations: (i) PHATE and persistent homology depend on design choices (distance, scale, filtration), which may affect effect sizes; (ii) we use simple linear/shallow MLP models, aiding interpretability but likely underestimating peak performance; and (iii) AHI-stratified analyses are associative and should not be interpreted causally.

Future work will explore end-to-end integration of manifold and topological structure, and test generalization across datasets and embedding models.
We will extend the late-fusion MLP to a \emph{dynamic fusion} architecture that assigns context-dependent weights to branches (embeddings, PHATE, TDA, EHR), which allows the model to adaptively route information across modalities and could improve both calibration and robustness.

\section{Conclusion}
We investigated the diagnostic information contained in the \textit{sequences} of per-epoch PedSleepMAE embeddings, and presented a late-fusion pipeline that augments the embeddings with PHATE-based temporal descriptors, persistent-homology summaries of latent geometry, and structured EHR context. On 2.5k+ pediatric sleep studies, all three feature families showed AHI-stratified associations, with topology and movement metrics tracking disease severity and several EHR variables stratifying independently. 

In predictive tasks, adding context consistently improved calibration and interpretability over embedding-only baselines, with different modalities contributing in a task-specific manner. The full late-fusion model (M3) achieved the best calibration (lowest Brier/ECE) across all four binary tasks and the top AUPRC for EEG arousal (0.478), while desaturation peaked with Emb+TDA (M1.2; 0.343) and hypopnea with Emb+EHR+TDA (M2.1; 0.219); apnea favored Emb+EHR+PHATE (M2; 0.142).

Together, these findings suggest that latent trajectory, topology, and clinical context capture complementary dimensions of pediatric sleep.
More broadly, our framework provides a lens into the structure of generative embeddings, offering one way to interpret what multimodal sleep models encode, rather than merely how well they classify.

\section*{Acknowledgements}
The authors thank Saurav Raj Pandey for his help with the dataset and the Longleaf cluster. Additionally, the first author thanks Baiming Zou for early guidance on the paper’s direction, and the second author thanks Jeremy Purvis for suggesting PHATE.

\bibliography{jmlr-sample}

\newpage
\appendix
\makeatletter
\pretocmd{\section}{\FloatBarrier}{}{}
\makeatother
\twocolumn[{%
\section{UMAP vs. PHATE}\label{app:umap}

\begin{center}
  \includegraphics[width=\textwidth]{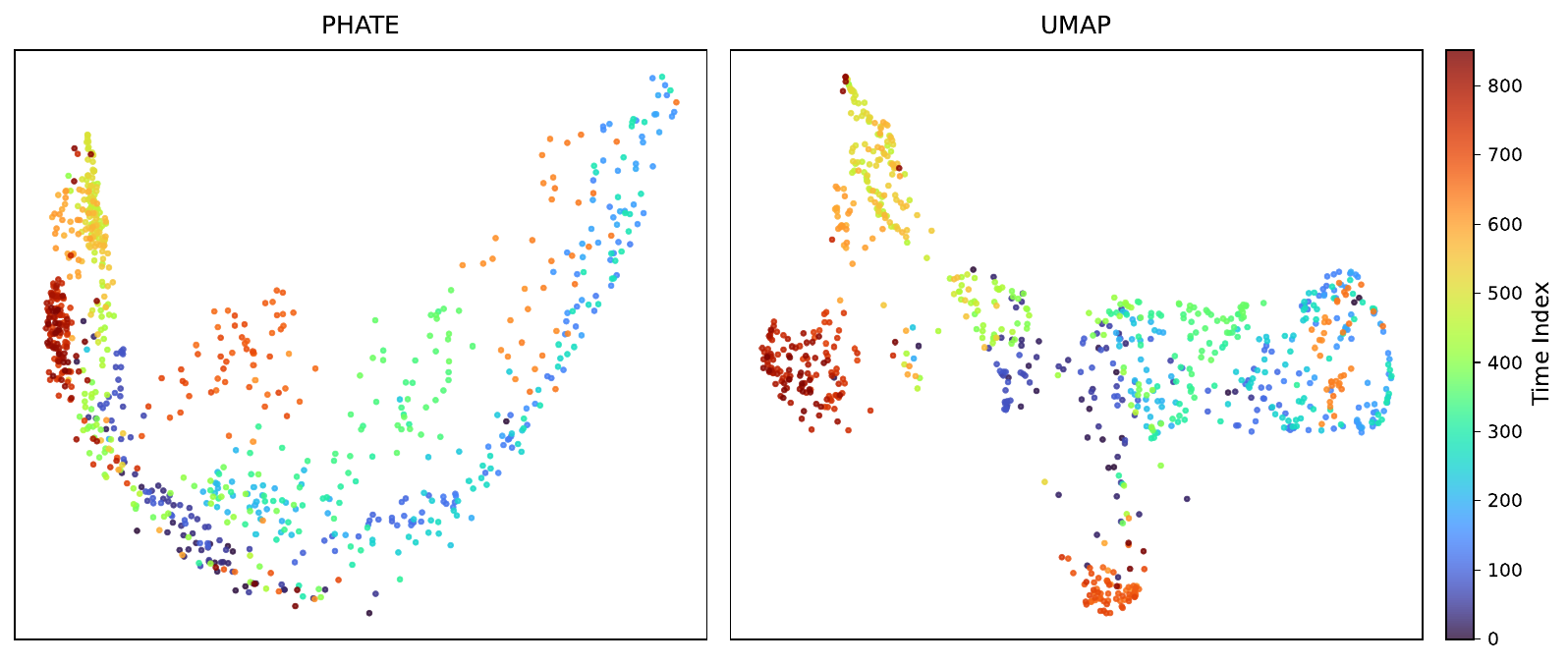}
  \captionof{figure}{Comparison of PHATE and UMAP embeddings for the same held-out PSG, colored by epoch index. PHATE reveals a smooth, time-continuous trajectory that preserves global temporal structure, whereas UMAP fragments the data into local clusters with weaker overall ordering.}
  \refstepcounter{figure}\label{fig:app-phate-umap}
\end{center}
\vspace{0.8em}

\begin{center}
\begin{minipage}{\textwidth}
\centering
\captionof{table}{Comparison of PHATE (M1.1) and UMAP on a single held-out session-wise split. Each manifold was fit on the training set and used to transform validation/test sessions.}
\refstepcounter{table}\label{tab:app-umap}
\begin{tabular}{@{}lcccc@{}}
\toprule
Task & Acc & F1 & ROC--AUC & AUPRC \\
\midrule
\multicolumn{5}{l}{\emph{Sleep staging (macro)}}\\
emb{+}PHATE (M1.1) & 0.708 & 0.675 & -- & -- \\
emb{+}UMAP  & 0.686 & 0.660 & -- & -- \\
\midrule
\multicolumn{5}{l}{\emph{Desaturation}}\\
emb{+}PHATE (M1.1) & 0.864 & 0.387 & 0.806 & 0.362 \\
emb{+}UMAP  & 0.863 & 0.339 & 0.779 & 0.346 \\
\midrule
\multicolumn{5}{l}{\emph{EEG arousal}}\\
emb{+}PHATE (M1.1) & 0.948 & 0.523 & 0.928 & 0.489 \\
emb{+}UMAP  & 0.949 & 0.486 & 0.916 & 0.445 \\
\midrule
\multicolumn{5}{l}{\emph{Apnea}}\\
emb{+}PHATE (M1.1) & 0.986 & 0.234 & 0.878 & 0.123 \\
emb{+}UMAP  & 0.984 & 0.186 & 0.863 & 0.106 \\
\midrule
\multicolumn{5}{l}{\emph{Hypopnea}}\\
emb{+}PHATE (M1.1) & 0.971 & 0.306 & 0.886 & 0.254 \\
emb{+}UMAP  & 0.955 & 0.204 & 0.828 & 0.226 \\
\bottomrule
\end{tabular}
\end{minipage}
\end{center}
\vspace{0.8em}
}]

\twocolumn[{%
\section{Prevalence}\label{app:prevalence}
\vspace{-0.6em}

Table~\ref{tab:prevalence} summarizes class distributions across train/validation/test splits.
Subgroup prevalence is included in Table~\ref{tab:app-prevalence-part1} to further highlight
age and demographic heterogeneity.

\begin{center}
\begin{minipage}{\textwidth}
\centering
\captionof{table}{Class distributions by split. Sleep staging shows \% for \{W, REM, N1, N2, N3\}; 
binary tasks list positive prevalence (\%).}
\label{tab:prevalence}
\begin{tabular}{@{}lp{4cm}p{4cm}p{4cm}@{}}
\toprule
Task & Train & Val & Test \\
\midrule
Sleep staging & [17.8, 3.6, 39.0, 22.5, 17.2] & [18.5, 3.7, 39.3, 22.1, 16.4] & [17.6, 3.7, 38.8, 23.0, 16.9] \\
Desaturation (+) & 8.738 & 8.843 & 9.419 \\
EEG arousal (+)  & 4.676 & 4.741 & 4.746 \\
Apnea (+)        & 0.846 & 0.622 & 0.808 \\
Hypopnea (+)     & 1.969 & 1.797 & 2.054 \\
\bottomrule
\end{tabular}
\end{minipage}
\end{center}

\vspace{1em}

\begin{center}
\begin{minipage}{\textwidth}
\centering
\captionof{table}{Subgroup prevalence and subgroup share of sessions for Apnea and Desaturation tasks.}
\label{tab:app-prevalence-part1}
\vspace{0.8em}
\begin{tabular}{@{}l l r r@{}}
\toprule
Task & Subgroup & \% positive (epochs) & \% of sessions \\
\midrule
Apnea & Infant/Toddler (0--1y) & 1.32 & 4.6 \\
      & Preschool (2--5y)      & 0.92 & 13.8 \\
      & Child (6--11y)         & 0.48 & 33.7 \\
      & Adolescent (12--17y)   & 0.95 & 37.5 \\
      & Adult (18+)            & 0.65 & 10.4 \\
      & Asian                  & 0.84 & 16.9 \\
      & Black                  & 1.18 & 7.2 \\
      & White                  & 0.71 & 63.8 \\
      & Multiracial            & 0.74 & 4.1 \\
      & Other Race             & 0.08 & 1.0 \\
      & Unknown                & 1.09 & 7.0 \\
      & Male                   & 0.89 & 52.5 \\
      & Female                 & 0.74 & 46.9 \\
      & Other Gender           & 0.10 & 0.6 \\
\midrule
Desaturation & Infant/Toddler (0--1y) & 12.92 & 11.2 \\
             & Preschool (2--5y)      & 8.63  & 26.5 \\
             & Child (6--11y)         & 6.72  & 31.6 \\
             & Adolescent (12--17y)   & 9.61  & 25.4 \\
             & Adult (18+)            & 9.81  & 5.3 \\
             & Asian                  & 9.57  & 2.9 \\
             & Black                  & 9.15  & 20.8 \\
             & White                  & 8.73  & 65.0 \\
             & Multiracial            & 8.67  & 8.0 \\
             & Other Race             & 3.78  & 0.2 \\
             & Unknown                & 8.28  & 3.2 \\
             & Male                   & 9.27  & 56.9 \\
             & Female                 & 8.20  & 43.1 \\
             & Other Gender           & 24.53 & 0.0 \\
\bottomrule
\end{tabular}
\end{minipage}
\end{center}

\vspace{0.6em}
}]

\begin{table*}[t]
\centering
\caption{Subgroup prevalence and subgroup share of sessions for EEG Arousal and Hypopnea tasks.}
\vspace{0.8em}
\label{tab:app-prevalence-part2}
\begin{tabular}{@{}l l r r@{}}
\toprule
Task & Subgroup & \% positive (epochs) & \% of sessions \\
\midrule
EEG arousal  & Infant/Toddler (0--1y) & 4.88 & 11.2 \\
             & Preschool (2--5y)      & 4.96 & 26.5 \\
             & Child (6--11y)         & 4.53 & 31.6 \\
             & Adolescent (12--17y)   & 4.48 & 25.4 \\
             & Adult (18+)            & 4.82 & 5.3 \\
             & Asian                  & 4.36 & 2.9 \\
             & Black                  & 4.76 & 20.8 \\
             & White                  & 4.69 & 65.0 \\
             & Multiracial            & 4.49 & 8.0 \\
             & Other Race             & 3.56 & 0.2 \\
             & Unknown                & 5.15 & 3.2 \\
             & Male                   & 4.79 & 56.9 \\
             & Female                 & 4.55 & 43.1 \\
             & Other Gender           & 3.52 & 0.0 \\
\midrule
Hypopnea     & Infant/Toddler (0--1y) & 1.81 & 11.2 \\
             & Preschool (2--5y)      & 1.63 & 26.5 \\
             & Child (6--11y)         & 1.49 & 31.6 \\
             & Adolescent (12--17y)   & 2.76 & 25.4 \\
             & Adult (18+)            & 3.17 & 5.3 \\
             & Asian                  & 2.31 & 2.9 \\
             & Black                  & 2.42 & 20.8 \\
             & White                  & 1.86 & 65.0 \\
             & Multiracial            & 1.62 & 8.0 \\
             & Other Race             & 0.25 & 0.2 \\
             & Unknown                & 1.68 & 3.2 \\
             & Male                   & 2.10 & 56.9 \\
             & Female                 & 1.78 & 43.1 \\
             & Other Gender           & 1.45 & 0.0 \\
\bottomrule
\end{tabular}
\end{table*}

\clearpage
\section{Implementation Details} \label{app:implementation}

\paragraph{Branch encoders.}
Each modality is encoded separately by a shallow block: a linear layer mapping the raw input dimension to 128 units, followed by layer normalization and ReLU activation:
\[
z^{(k)} = \mathrm{ReLU}\!\Big(\mathrm{LN}\!\big(W^{(k)} x^{(k)} + b^{(k)}\big)\Big), \qquad z^{(k)} \in \mathbb{R}^{128}.
\]
The five possible inputs are: embeddings (7{,}680-D), EHR (23-D), PHATE-point (6-D), PHATE-time (6-D), and TDA (6-D). Each branch yields a 128-D latent representation.

\paragraph{Fusion and classifier head.}
Encoded features are concatenated into a single latent vector. Depending on branch inventory, this vector ranges from 128-D (M0.1) to 640-D (M3). It is passed through a two-layer classifier head: Linear $\to$ 256 units, ReLU, Dropout(0.30), and Linear $\to K$ logits. Thus each model differs only in which branches are included; the head is otherwise matched across ablations.

\paragraph{Training protocol.}
We used subject-wise splits to avoid data leakage across sessions from the same participant. All models were trained with AdamW (learning rate $10^{-3}$, weight decay $10^{-5}$), batch size~256, and automatic mixed precision. 
Learning rate decay was controlled by a \texttt{ReduceLROnPlateau} scheduler (factor~0.5, patience~3), and early stopping after 8~non-improving validations. We performed 5-fold subject-wise cross-validation using the same training recipe and report mean~±~SD across folds. 
Binary decision thresholds were chosen by maximizing validation~F1; multiclass sleep staging reported macro–F1.

\paragraph{Normalization.}
All features were standardized with train-only statistics. For stability, values were clipped to $[-8,8]$. Epoch-level features (embeddings, PHATE-point) were normalized across all epochs, while session-level vectors (EHR, PHATE-time, TDA) were normalized across sessions and broadcast to all epochs. Non-finite values were set to zero before normalization.

\paragraph{Loss functions and imbalance handling.}
To handle severe class imbalance, we applied class-weighted losses with $w_k \propto 1/\text{freq}_k$. Binary tasks used focal cross-entropy with focusing parameter $\gamma=1.5$ \citep{lin2017focal}, while sleep staging used weighted cross-entropy.

\paragraph{Architecture schematic.}
For reference, Figure~\ref{fig:latefusion-mlp} (in the Appendix) visualizes the full M3 late-fusion MLP.

\begin{figure*}[t]  
  \centering
  \includegraphics[width=\textwidth]{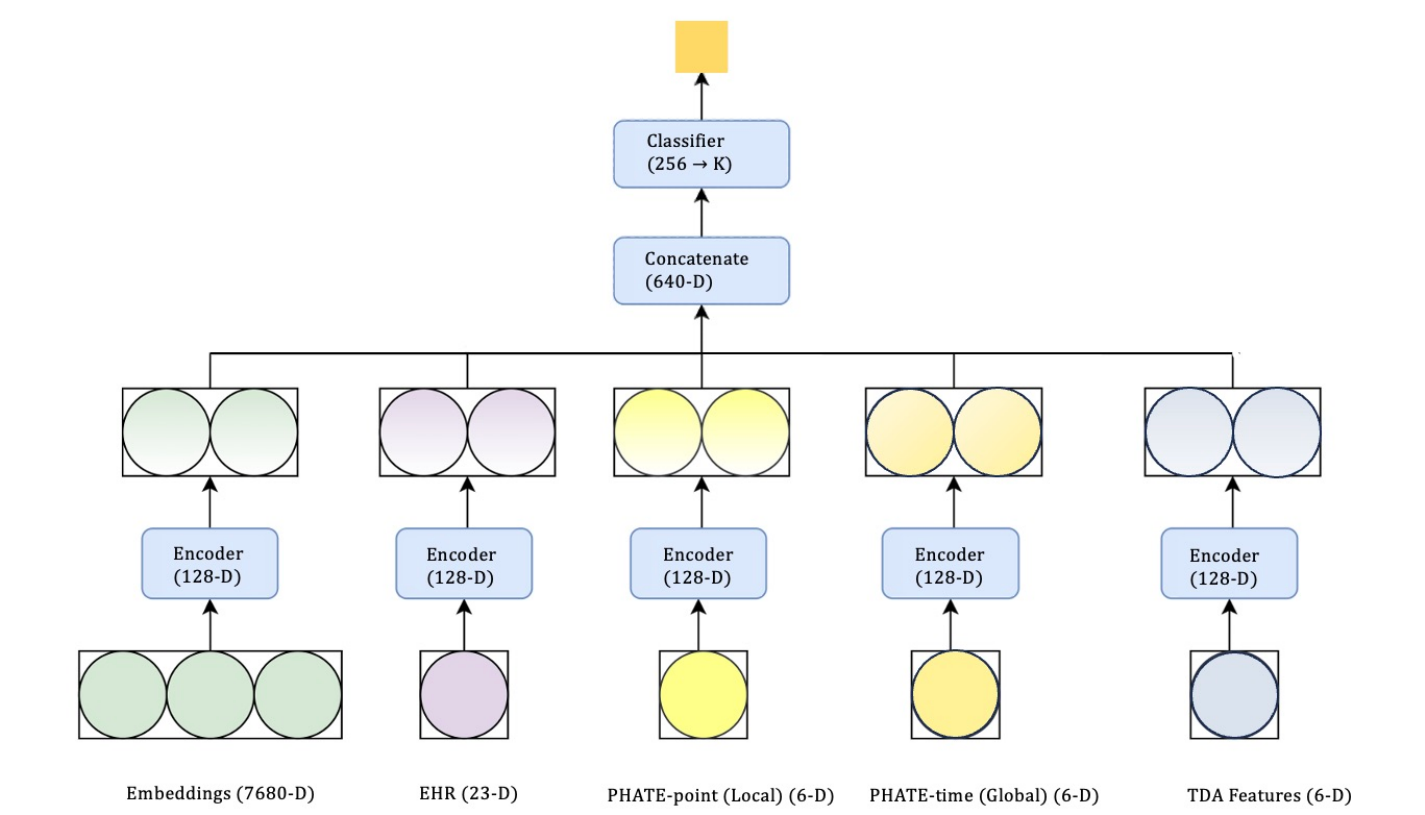} 
  \caption{Late-fusion MLP (M3). Each branch input is encoded with Linear→LayerNorm→ReLU (128-D).
  Latents are concatenated (5×128=640-D) and passed through a classifier head: Linear 640→256, ReLU,
  Dropout(0.30), and Linear 256→K. Per-epoch branches are Embeddings and PHATE-point; session-level
  branches are EHR, PHATE-time, and TDA (broadcast across epochs).}
  \label{fig:latefusion-mlp}
\end{figure*}

\clearpage

\twocolumn[{%
\section{AHI Tests}\label{sec:app_ahi}
\vspace{-0.6em}

\begin{center}
\begin{minipage}{\textwidth}
\centering
\captionof{table}{Permutation Kruskal--Wallis omnibus tests across AHI groups for 
pre-specified session-level descriptors (Table~\ref{tab:branch-inventory}) (significant results only; \(q<0.05\)). 
Larger $H$ (with small $q$ after Holm correction) indicates stronger distributional differences across AHI strata.}
\label{tab:ahi-omnibus}
\vspace{0.4em}
\begin{tabular}{@{}lcc@{}}
\toprule
Feature & $H$ & $q$ \\
\midrule
H0\_n\_bars        & 54.095 & 0.002 \\
H1\_n\_bars        & 23.266 & 0.002 \\
H1\_max\_pers      & 13.174 & 0.002 \\
Betti--L$^{2}$     & 29.032 & 0.002 \\
ratio\_sum\_H1\_H0 & 21.960 & 0.002 \\
H0\_sum\_pers      & 12.235 & 0.009 \\
mean step          & 16.179 & 0.006 \\
max step           & 19.217 & 0.002 \\
mean turn          & 17.081 & 0.002 \\
\bottomrule
\end{tabular}
\end{minipage}
\end{center}

\vspace{0.8em} 

\begin{center}
\begin{minipage}{\textwidth}
\centering
\captionof{table}{Permutation Mann–Whitney contrasts for AHI extremes on pre-specified descriptors
(significant results only; \(q<0.05\)). Effect is median(group)$-$median(others); 
$\delta$ is Cliff’s delta.}
\label{tab:ahi-contrasts}
\vspace{0.4em}
\begin{tabular}{@{}lccc c@{}}
\toprule
Feature & Effect ($\Delta$) & $\delta$ & $q$ & Direction \\
\midrule
\multicolumn{5}{l}{\textit{Severe (AHI$\ge$10) vs all others}} \\
\midrule
H0\_n\_bars        & -16.500 & -0.227 & 0.003 & lower \\
H1\_n\_bars        & -20.500 & -0.168 & 0.003 & lower \\
Betti--L$^{2}$     & -23.660 & -0.166 & 0.003 & lower \\
ratio\_sum\_H1\_H0 & -0.004  & -0.153 & 0.003 & lower \\
H1\_max\_pers      & 0.197   & 0.108  & 0.010 & higher \\
H0\_sum\_pers      & 33.510  & 0.087  & 0.025 & higher \\
mean step          & 0.104   & 0.137  & 0.003 & higher \\
\midrule
\multicolumn{5}{l}{\textit{Healthy (AHI$<$1) vs all others}} \\
\midrule
H0\_n\_bars        & 8.000   & 0.107  & 0.005 & higher \\
Betti--L$^{2}$     & 10.640  & 0.089  & 0.005 & higher \\
H0\_sum\_pers      & -22.250 & -0.070 & 0.005 & lower \\
ratio\_sum\_H1\_H0 & 0.003   & 0.076  & 0.005 & higher \\
H1\_max\_pers      & -0.067  & -0.065 & 0.017 & lower \\
H1\_n\_bars        & 5.000   & 0.059  & 0.019 & higher \\
mean step          & -0.018  & -0.058 & 0.024 & lower \\
\bottomrule
\end{tabular}
\end{minipage}
\end{center}

\vspace{0.6em}
}]  

\begin{figure*}[t]
\centering
\begin{minipage}{0.48\linewidth}
\centering
\captionof{table}{Permutation Kruskal--Wallis omnibus tests across AHI groups for all 23 EHR features.
Larger $H$ (with small $q$ after Holm correction) indicates stronger distributional differences across AHI strata.}
\label{tab:ahi-ehr-omnibus}
\vspace{0.4em}
\begin{tabular}{@{}lcc@{}}
\toprule
EHR Feature & $H$ & $q$ \\
\midrule
Age (z)                   & 10.629 & 0.016 \\
Gender (male)             & 15.669 & 0.003 \\
Gender (female)           & 3.242  & 0.214 \\
Gender (other/unknown)    & 0.821  & 0.744 \\
Race (White)              & 11.140 & 0.013 \\
Race (Black)              & 25.045 & 0.001 \\
Race (Asian)              & 2.512  & 0.332 \\
Race (Other)              & 1.921  & 0.411 \\
Race (Missing/unknown)    & 0.873  & 0.691 \\
Ethnicity (Hispanic)      & 1.002  & 0.602 \\
Asthma                    & 4.011  & 0.131 \\
Obesity                   & 91.098 & 0.001 \\
Diabetes                  & 24.995 & 0.000 \\
Hypertension              & 23.704 & 0.001 \\
Depression/mood disorder  & 9.851  & 0.018 \\
Anxiety                   & 20.689 & 0.001 \\
ADHD                      & 2.144  & 0.299 \\
Seizure disorder/epilepsy & 3.821  & 0.145 \\
GERD                      & 8.484  & 0.041 \\
Cerebral palsy            & 1.301  & 0.515 \\
Autism                    & 2.015  & 0.368 \\
Developmental delay       & 1.774  & 0.447 \\
\bottomrule
\end{tabular}
\end{minipage}\hfill
\begin{minipage}{0.48\linewidth}
\centering
\captionof{table}{Permutation Mann--Whitney contrasts for AHI extremes on EHR features
(significant results only; \(q<0.05\)). 
Effect (\(\Delta\)) = median(group)$-$median(others) are small and omitted for brevity; 
$\delta$ is Cliff’s delta.}
\label{tab:ehr-contrasts}
\vspace{0.4em}
\begin{tabular}{@{}lccc@{}}
\toprule
Feature & $\delta$ & $q$ & Direction \\
\midrule
\multicolumn{4}{l}{\textit{Healthy (AHI$<$1) vs all others}} \\
\midrule
Obesity & +0.180 & 0.001 & higher \\
Diabetes & +0.125 & 0.001 & higher \\
Hypertension & +0.110 & 0.001 & higher \\
ADHD & +0.115 & 0.001 & higher \\
Sex: Female & +0.090 & 0.003 & higher \\
Sex: Male & -0.085 & 0.003 & lower \\
Depression/Mood & +0.070 & 0.008 & higher \\
Race: Black & -0.060 & 0.030 & lower \\
Race: White & +0.058 & 0.030 & higher \\
Race: Asian & -0.050 & 0.045 & lower \\
Ethnicity: Hispanic & +0.052 & 0.030 & higher \\
\midrule
\multicolumn{4}{l}{\textit{Mild (1$\le$AHI$<$5) vs all others}} \\
\midrule
ADHD & +0.095 & 0.010 & higher \\
Age (z) & +0.070 & 0.031 & higher \\
\midrule
\multicolumn{4}{l}{\textit{Moderate (5$\le$AHI$<$10) vs all others}} \\
\midrule
Obesity & +0.100 & 0.005 & higher \\
\midrule
\multicolumn{4}{l}{\textit{Severe (AHI$\ge$10) vs all others}} \\
\midrule
Hypertension & +0.098 & 0.004 & higher \\
Race: Black & +0.120 & 0.002 & higher \\
Race: White & -0.080 & 0.015 & lower \\
\bottomrule
\end{tabular}
\end{minipage}
\end{figure*}
\FloatBarrier

\twocolumn[{%
\section{Additional Predictive Results}\label{app:results}
\vspace{-0.8em}

\begin{center}
  \includegraphics[width=\textwidth]{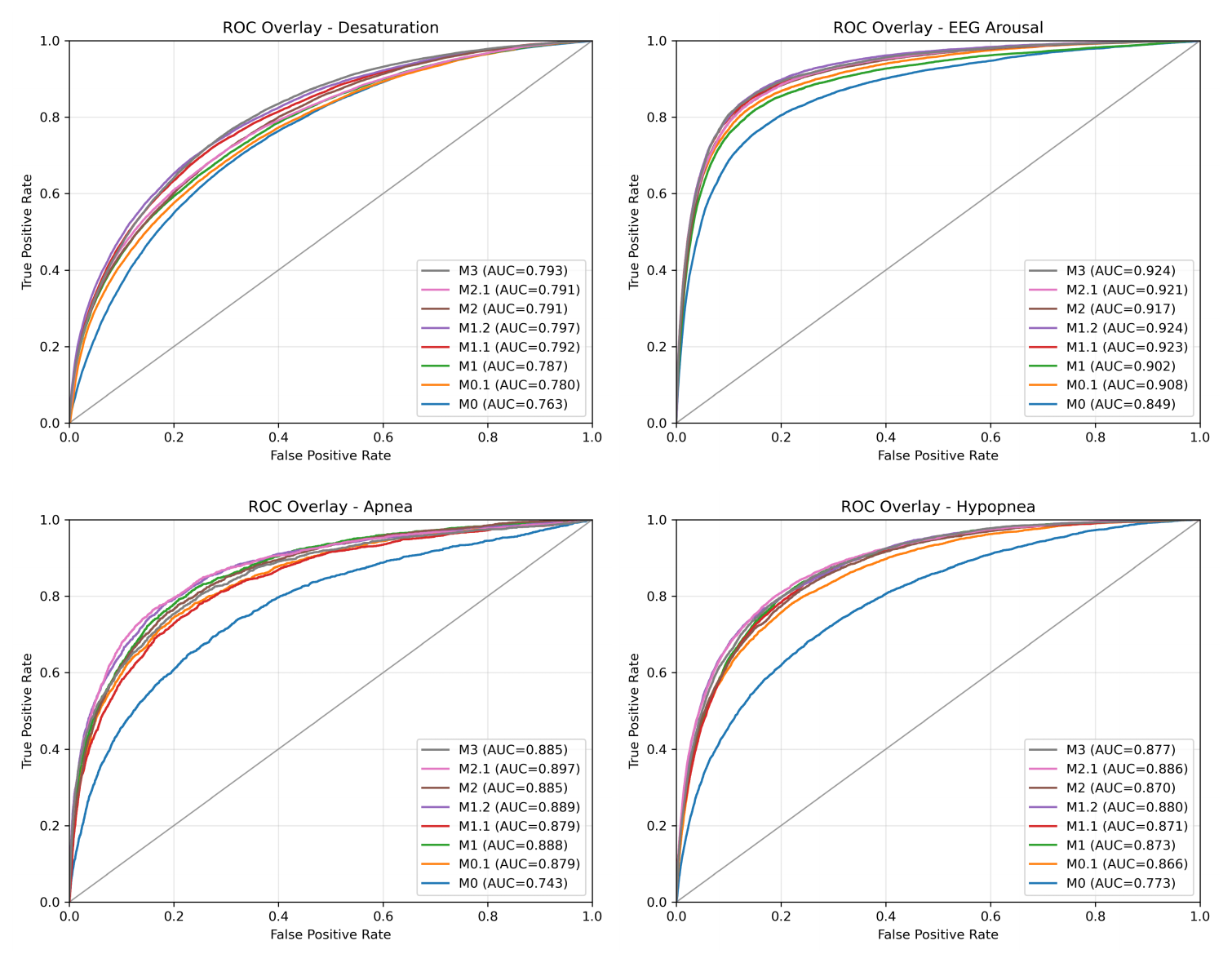}
  \captionof{figure}{ROC curves for the four binary tasks on the test set.
  Each panel overlays all ablation models, with legend reporting ROC–AUC.
  These curves complement the precision–recall plots in Figure~\ref{fig:3} and
  illustrate consistent improvements beyond the linear probe baseline once
  contextual branches are added.}
  \label{fig:roc_curves}
\end{center}
\vspace{0.8em}
}]
\clearpage

\section{Full Test-Set Metrics}
\label{app:metrics}

\noindent \textit{Reading guide.}
Below are the full test-set results with 5-fold subject-wise cross-validation. Table~\ref{tab:staging} reports multiclass sleep staging results (balanced accuracy and macro–F1 as mean ± SD). 
Tables~\ref{tab:desat}–\ref{tab:hypop} list Accuracy, F1, ROC–AUC, AUPRC (reported as mean ± SD), and the calibration metrics 
Brier score and ECE for the four binary tasks, following the same ablation order. 
Lower Brier and ECE values indicate better-calibrated probability estimates, which are 
especially important for clinical interpretability.

\begin{table}[htbp]
\centering
\caption{Multiclass Sleep staging.}
\vspace{0.8em}
\label{tab:staging}
\resizebox{0.9\linewidth}{!}{
\begin{tabular}{lcc}
\toprule
Model & Balanced Accuracy & Macro–F1 \\
\midrule
M0 (linear, emb)        & 0.726 ± 0.002 & 0.655 ± 0.001 \\
M0.1 (mlp, emb)         & 0.734 ± 0.010 & 0.664 ± 0.004 \\
M1 (emb+EHR)            & 0.739 ± 0.005 & 0.664 ± 0.004 \\
M1.1 (emb+PHATE)        & 0.743 ± 0.009 & 0.672 ± 0.004 \\
M1.2 (emb+TDA)          & 0.740 ± 0.006 & 0.671 ± 0.005 \\
M2 (emb+EHR+PHATE)      & 0.756 ± 0.002 & 0.679 ± 0.003 \\
M2.1 (emb+EHR+TDA)      & 0.749 ± 0.004 & 0.671 ± 0.003 \\
M3 (all)                & 0.762 ± 0.004 & 0.678 ± 0.004 \\
\bottomrule
\end{tabular}}
\end{table}
\FloatBarrier

\begin{table}[htbp]
\centering
\caption{Desaturation.}
\vspace{0.8em}
\label{tab:desat}
\resizebox{0.98\linewidth}{!}{
\begin{tabular}{lcccccc}
\toprule
Model & Accuracy & F1 & ROC–AUC & AUPRC & Brier & ECE \\
\midrule
M0   & 0.833 ± 0.020 & 0.324 ± 0.018 & 0.763 ± 0.019 & 0.257 ± 0.022 & 0.194 & 0.293 \\
M0.1 & 0.870 ± 0.011 & 0.348 ± 0.012 & 0.780 ± 0.004 & 0.314 ± 0.021 & 0.157 & 0.231 \\
M1   & 0.870 ± 0.006 & 0.365 ± 0.010 & 0.787 ± 0.004 & 0.337 ± 0.013 & 0.147 & 0.194 \\
M1.1 & 0.867 ± 0.007 & 0.368 ± 0.012 & 0.792 ± 0.005 & 0.335 ± 0.017 & 0.155 & 0.209 \\
M1.2 & 0.875 ± 0.010 & 0.374 ± 0.019 & 0.797 ± 0.007 & 0.343 ± 0.028 & 0.161 & 0.243 \\
M2   & 0.887 ± 0.012 & 0.366 ± 0.013 & 0.791 ± 0.007 & 0.342 ± 0.018 & 0.144 & 0.184 \\
M2.1 & 0.889 ± 0.006 & 0.371 ± 0.020 & 0.791 ± 0.006 & 0.338 ± 0.027 & 0.142 & 0.187 \\
M3   & 0.889 ± 0.006 & 0.373 ± 0.014 & 0.793 ± 0.003 & 0.342 ± 0.022 & 0.138 & 0.183 \\
\bottomrule
\end{tabular}}
\end{table}
\FloatBarrier

\begin{table}[htbp]
\centering
\caption{EEG arousal.}
\vspace{0.8em}
\label{tab:eeg}
\resizebox{0.98\linewidth}{!}{
\begin{tabular}{lcccccc}
\toprule
Model & Accuracy & F1 & ROC–AUC & AUPRC & Brier & ECE \\
\midrule
M0   & 0.930 ± 0.005 & 0.393 ± 0.014 & 0.849 ± 0.011 & 0.309 ± 0.013 & 0.129 & 0.225 \\
M0.1 & 0.944 ± 0.003 & 0.420 ± 0.007 & 0.908 ± 0.004 & 0.454 ± 0.007 & 0.104 & 0.168 \\
M1   & 0.945 ± 0.001 & 0.481 ± 0.006 & 0.902 ± 0.005 & 0.437 ± 0.014 & 0.092 & 0.140 \\
M1.1 & 0.950 ± 0.003 & 0.508 ± 0.008 & 0.923 ± 0.004 & 0.476 ± 0.015 & 0.092 & 0.149 \\
M1.2 & 0.947 ± 0.004 & 0.506 ± 0.009 & 0.924 ± 0.003 & 0.475 ± 0.013 & 0.094 & 0.151 \\
M2   & 0.949 ± 0.002 & 0.495 ± 0.010 & 0.917 ± 0.006 & 0.460 ± 0.012 & 0.100 & 0.164 \\
M2.1 & 0.949 ± 0.002 & 0.497 ± 0.013 & 0.921 ± 0.005 & 0.468 ± 0.017 & 0.098 & 0.158 \\
M3   & 0.949 ± 0.002 & 0.508 ± 0.011 & 0.924 ± 0.005 & 0.478 ± 0.014 & 0.084 & 0.130 \\
\bottomrule
\end{tabular}}
\end{table}
\FloatBarrier

\begin{table}[htbp]
\centering
\caption{Apnea.}
\vspace{0.8em}
\label{tab:apnea}
\resizebox{0.98\linewidth}{!}{
\begin{tabular}{lcccccc}
\toprule
Model & Accuracy & F1 & ROC–AUC & AUPRC & Brier & ECE \\
\midrule
M0   & 0.976 ± 0.009 & 0.105 ± 0.029 & 0.743 ± 0.050 & 0.043 ± 0.013 & 0.055 & 0.089 \\
M0.1 & 0.982 ± 0.001 & 0.179 ± 0.023 & 0.879 ± 0.013 & 0.092 ± 0.015 & 0.066 & 0.133 \\
M1   & 0.985 ± 0.001 & 0.210 ± 0.033 & 0.888 ± 0.013 & 0.129 ± 0.036 & 0.032 & 0.054 \\
M1.1 & 0.983 ± 0.001 & 0.199 ± 0.015 & 0.879 ± 0.008 & 0.107 ± 0.009 & 0.038 & 0.051 \\
M1.2 & 0.985 ± 0.004 & 0.204 ± 0.019 & 0.889 ± 0.016 & 0.122 ± 0.019 & 0.037 & 0.058 \\
M2   & 0.985 ± 0.003 & 0.231 ± 0.042 & 0.885 ± 0.014 & 0.142 ± 0.029 & 0.030 & 0.045 \\
M2.1 & 0.988 ± 0.002 & 0.220 ± 0.035 & 0.897 ± 0.015 & 0.141 ± 0.036 & 0.027 & 0.049 \\
M3   & 0.982 ± 0.006 & 0.219 ± 0.038 & 0.885 ± 0.023 & 0.136 ± 0.036 & 0.014 & 0.016 \\
\bottomrule
\end{tabular}}
\end{table}
\FloatBarrier

\begin{table}[htbp]
\centering
\caption{Hypopnea.}
\vspace{0.8em}
\label{tab:hypop}
\resizebox{0.98\linewidth}{!}{
\begin{tabular}{lcccccc}
\toprule
Model & Accuracy & F1 & ROC–AUC & AUPRC & Brier & ECE \\
\midrule
M0   & 0.950 ± 0.007 & 0.167 ± 0.015 & 0.773 ± 0.010 & 0.088 ± 0.011 & 0.158 & 0.241 \\
M0.1 & 0.958 ± 0.003 & 0.238 ± 0.025 & 0.866 ± 0.008 & 0.155 ± 0.024 & 0.101 & 0.170 \\
M1   & 0.965 ± 0.004 & 0.257 ± 0.036 & 0.873 ± 0.011 & 0.193 ± 0.027 & 0.092 & 0.146 \\
M1.1 & 0.963 ± 0.004 & 0.273 ± 0.036 & 0.871 ± 0.015 & 0.185 ± 0.026 & 0.080 & 0.128 \\
M1.2 & 0.962 ± 0.006 & 0.283 ± 0.034 & 0.880 ± 0.012 & 0.195 ± 0.026 & 0.089 & 0.151 \\
M2   & 0.968 ± 0.006 & 0.267 ± 0.027 & 0.870 ± 0.012 & 0.197 ± 0.020 & 0.077 & 0.122 \\
M2.1 & 0.966 ± 0.006 & 0.309 ± 0.021 & 0.886 ± 0.007 & 0.219 ± 0.020 & 0.086 & 0.134 \\
M3   & 0.965 ± 0.005 & 0.269 ± 0.029 & 0.877 ± 0.010 & 0.186 ± 0.022 & 0.067 & 0.103 \\
\bottomrule
\end{tabular}}
\end{table}
\FloatBarrier

\Needspace{10\baselineskip}

\end{document}